%% file: Main_text.tex
\pdfoutput=1

\documentclass[lettersize,journal]{IEEEtran}
\usepackage{amsmath,amsfonts}
\usepackage{algorithmic}
\usepackage{algorithm}
\usepackage{array}
\usepackage[caption=false,font=normalsize,labelfont=sf,textfont=sf]{subfig}
\usepackage{textcomp}
\usepackage{stfloats}
\usepackage{url}
\usepackage{verbatim}
\usepackage{graphicx}
\usepackage{cite}
\usepackage{setspace}
\usepackage{multirow}
\usepackage[justification=centering]{caption}

\begin{document}
\title{Large Scale Foundation Models for Intelligent Manufacturing Applications: A Survey}
\author{Haotian Zhang, Semujju Stuart Dereck, Zhicheng Wang, Xianwei Lv, Kang Xu, Liang Wu, Ye Jia, Jing Wu, Zhuo Long, Wensheng Liang, X.G. Ma \IEEEmembership{Member, IEEE}, and Ruiyan Zhuang 
\thanks{This work was supported in part by the Guangdong Basic and Applied Basic Research Fund Project under Grant 2022A1515140121; and in part by the Initiation Funding of Foshan Graduate School of Innovation, Northeastern University under Grant 200076421002 }
\thanks{Haotian Zhang, Zhicheng Wang, Xianwei Lv, Kang Xu, Liang Wu, and X. Ma are with the College of Information Science and Engineering, Northeastern University, Shenyang 110819, China, and the Foshan Graduate School, Northeastern University, Foshan 528311, China. (E-mail:2200966@stu.neu.edu.cn, 2390108@stu.neu.edu.cn,2270888@stu.neu.edu.cn, maxg@mail.neu.edu.cn).}
\thanks{Wensheng Liang is with the School of mechanical engineering and automation, South China University of Technology, Shenyang 110819, China. (E-mail:2200385@stu.neu.edu.cn).}
\thanks{Semujju Stuart Dereck is with the School of Software Engineering, South China University of Technology, Guangzhou 510006, China. (E-mail:stuartsemujju@gmail.com).}
\thanks{Jing Wu is with the Departement of Computer and Information Science, Qinghai University of Science and Technology, Xining, China. (E-mail:mirror\_a3@hotmail.com).}
\thanks{Ruiyan Zhuang is with the Enterprise AI, Midea Group, Foshan 528311, China. (E-mail:zhuangry@midea.com).}
}

\maketitle

\begin{abstract}
Although the applications of artificial intelligence especially deep learning had greatly improved various aspects of intelligent manufacturing, they still face challenges for wide employment due to the poor generalization ability, difficulties to establish high-quality training datasets, and unsatisfactory performance of deep learning methods. The emergence of large scale foundational models(LSFMs) had triggered a wave in the field of artificial intelligence, shifting deep learning models from single-task, single-modal, limited data patterns to a paradigm encompassing diverse tasks, multimodal, and pre-training on massive datasets. Although LSFMs had demonstrated powerful generalization capabilities, automatic high-quality training dataset generation and superior performance across various domains, applications of LSFMs on intelligent manufacturing were still in their nascent stage. A systematic overview of this topic was lacking, especially regarding which challenges of deep learning can be addressed by LSFMs and how these challenges can be systematically tackled. To fill this gap, this paper systematically expounded current statue of LSFMs and their advantages in the context of intelligent manufacturing. and compared comprehensively with the challenges faced by current deep learning models in various intelligent manufacturing applications. We also outlined the roadmaps for utilizing LSFMs to address these challenges. Finally, case studies of applications of LSFMs in real-world intelligent manufacturing scenarios were presented to illustrate how LSFMs could help industries, improve their efficiency.
\end{abstract}



\input{Main_text/I_Introduction}

\input{Main_text/II_Previous_works_and_challenges_of_Deep_Learning_in_Intelligent_Manufacturing}

\input{Main_text/III_Background_for_LSFMs}

\input{Main_text/IV_Roadmaps_for_solving_problems_with_LSFMs}

\input{Main_text/V_Cases_of_application_of_LSFMs_in_intelligent_manufacturing}

\input{Main_text/VI_Conclusion}

\section*{Acknowledgment}
We would like to thank to Midea Group for providing us with data samples and a testing platform for our case study. We appreciate Zhenrui Wu for completing the drawing of Figure\ref{fig:LLM timeline} and collecting the references. We also thank Shuai Li, Wensheng Liang, Liu Junwei and Kang Xu for the technical support they provided for our case study.

\bibliographystyle{unsrt}
\bibliography{Large_Scale_Foundation_Models_for_Intelligent_Manufacturing_Applications__A_Survey}

\end{document}

%% file: Main_text/I_Introduction.tex
\section{Introduction}
\label{sec:introduction}

\IEEEPARstart{M}{anufacturing} industries are mainstays of a nation's economy, and several countries had announced strategic roadmaps to promote applications of new techniques of manufacturing to ensure their leadership in this area, e.g., Industry 4.0 of Germany\cite{rojko2017industry}, the Smart Manufacturing Leadership Coalition (SMLC) in the U.S\cite{bryner2012smart}, and China Manufacturing 2025\cite{wubbeke2016made}. Over the past several decades, manufacturing was getting more intelligent by deploying new technologies such as sensors, internet of things (loT), robotics, digital twins, and cyber-physical systems (CPSs)\cite{kumar2018methods,schutze2018sensors,gupta2022industrial,lampropoulos2018internet,lu2016internet,tantawi2019advances,bhatt2020expanding,goel2020robotics,kritzinger2018digital,lu2020digital,lee2015cyber,wang2023data}, wherein unprecedented amount of data were continuously generated and captured at all stages of manufacturing processes. Therefore, optimal data-processing algorithms were highly desired to realize efficient fault diagnosis and predictive maintenance, quality control, human operations, process optimization, and many other smart decision-makings needed for intelligent manufacturing\cite{akter2021algorithmic,dagli2012artificial,li2017applications,wang2022big,zhang2019big}. Statistics showed that 82\% of the industrial activities using intelligent manufacturing technologies obtained improved efficiency and performance\cite{coalitionmanufacturing,akter2021algorithmic}

These improvements of intelligent manufacturing largely attributed to the implementations of various machine learning algorithms with increasing scale and complexities of manufacturing data, wherein many advanced data-driven methods had been studied and employed to enable large-scale data processing capabilities with high efficiency and powerful decision-making abilities for highly non-linear optimizations, both of which were commonly required in sophisticated manufacturing activities. Table 1 listed some review papers in this area\cite{gan2020prognostics,singh2020systematic,polverino2023machine,weikun2023physics,wang2017prognostics,tsui2015prognostics,souza2023machine,sutharssan2015prognostic,dogan2021machine,coulter2018intelligent,zhang2023top,fahle2020systematic,rezaeianjouybari2020deep,zhang2019review,hu2022prognostics,ellefsen2019comprehensive,zhang2022prognostics,kumar2023prognostics,remadna2018overview,thoppil2021deep,jayalaxmi2022machine,xu2022review,jamwal2022deep,wang2018deep,malhan2023role,qin2022research,tran2022machine,he2023research,rahman2023machine,rajesh2022smart,deng2023research,wu2022progress,zhao2021applications,di2023review,neupane2020bearing,ming2021application,scibilia2023modeling,ramasamy2018recent,qiu2022multi,nguyen2021trends,kulsoom2022review,attal2015physical,lara2012survey,chen2021deep,gu2021survey,zhang2022deep,nweke2018deep,ramanujam2021human,li2019survey,kumar2021human,khan2021survey,danthala2018robotic,liu2022robot,wang2019artificial,semeraro2023human,gulzow2020recent,sajwan2023review,domae2019recent,aggarwal2022deep,soori2023artificial,de2021robotic,hernavs2018deep,valaskova2021deep,cordeiro2022bin,caldera2018review,duan2021robotics}.

\begin{figure*}
    \centering
    \includegraphics[width=1\textwidth]{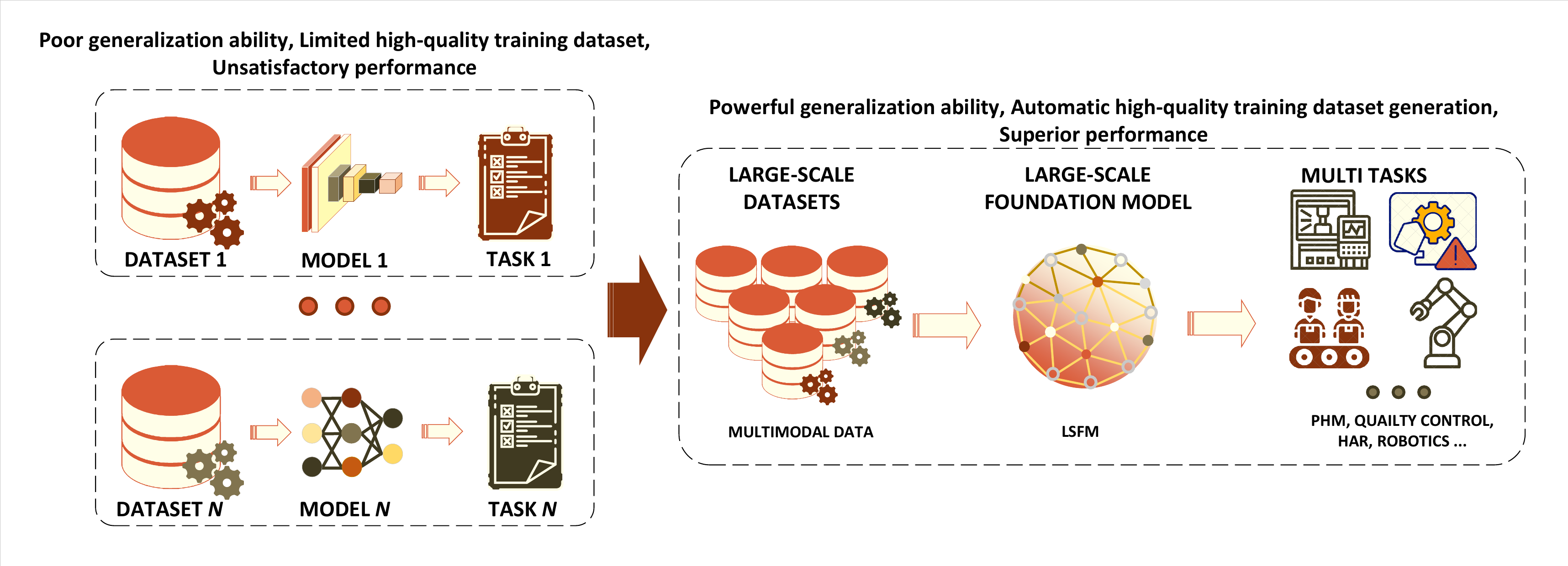}
    \caption{The evolution from deep learning to large scale foundation models(LSFMs)}
    \label{fig:overview}
\end{figure*}

\input{Table/ML_DL_survey_list}

Traditional machine learning methods, such as support vector machines, K-nearest neighbors. Naive Bayes, etc., could improve performance of decision-making\cite{rawat2021application,meyer2014machine,rai2021machine}, production line schedule\cite{li2020machine,takeda2020machine}, machine maintenance arrangements\cite{coraddu2016machine,paolanti2018machine}, failure prediction\cite{campos2018exploratory,celikmih2020failure,orru2020machine}, quality assessment\cite{xu2015visual,pastor2021quality}, and defect detection\cite{huang2018study,zhang2021learningadd} in manufacturing to certain extents. However, they relied heavily on handcrafting feature engineering to represent data with domain knowledge and were lack of capabilities to handle highly non-linear relationships among large-scale data, limiting their applications in intelligent manufacturing\cite{jogin2018feature,farias2016automatic}.

Deep learning, as one of the advanced machine learning methods, could enable automatic feature extractions and pattern recognitions from high-dimensional and nonlinear raw data by employing multi-level neural network architectures, making them more adaptive to the complicated data treatments of intelligent manufacturing. ln the last decade, deep learning methods was the mainstream data-driven approaches used in various areas of intelligent manufacturing, such as health management (PHM)\cite{zonta2022predictive,al2019multimodal,du2017deeplog,lee2013recent,vogl2019review,malhi2011prognosis,zhao2017machine,zhao2017learning,wu2018remaining,malhotra2015long,wang2017deep,deutsch2017remaining,qiu2014ensemble,zhang2017resource,huang2018bearing,arias2021aircraft,verstraete2017deep,li2016gearbox,lei2016intelligent,fang2023you,kumar2019multi}, quality control\cite{manyar2022synthetic,manyar2023physics,jain2022synthetic,tabernik2020segmentation,he2019end,li2021defectnet,bhatt2021image,volkau2019detection,xu2020weakly,weimer2016design,chen2020solar,czimmermann2020visual,samet2016primer,wei2019research,yang2020using,ren2017generic,masci2012steel}, robotic\cite{gulde2018ropose,chen2018industrial,wu2022cascaded,park2022development,trinh2022dynamics,khatib2008torque,wang2022transformer}, and human activity recognition\cite{chen2021deep,pienaar2019human,qi2020smartphone,arifoglu2017activity,bashar2020smartphone,rashid2022ahar,al2022multi,nafea2022multi,andrade2022human,yang2022activity,imran2022harresnext,hussain2022vision,wensel2023vit,liu2023transtm,do2022multi,yang2023transformer,li2022real}.

Although deep learning showed high level abstract representation abilities of feature learning, with great end-to-end decision-making modeling capabilities and significantly, reduced need for human-labor, allowing it to greatly push the development of intelligent manufacturing, it still faced significant difficulties when being employed\cite{whang2023data,jamwal2022deep,gupta2020deep,armstrong2015we,ravikumar2022challenges}. Firstly, the performance of small customized deep learning models tailored for specific patterns and objectives was limited. These models suffered from issues like limited generalization, poor interpretability, vulnerability to attacks, and could not meet the requirements of enterprises for intelligent production and management, especially in complex tasks with diverse data\cite{yuan2022recent,rajeswaran2018learning,shi2020neural,liu2020logiqa,gao2020survey,zhang2019multimodal} . In addition, they could only handle individual tasks in a scattered and loose coupling manner\cite{wang2023production,ajakwe2022dronet,zhang2022machining}.

Secondly, growing requirements of data scale and dataset establishment cost constrained the performance of deep learning models. As a data-driven approach, deep learning models rely on fitting the relationships between inputs and outputs wherein the volume and quality of the training datasets play critical roles\cite{whang2023data}. Although the use of new technologies like sensors and the Internet of Things had made it possible to efficiently collect a massive amount of data\cite{de2020novel,liu2018blockchain,wang2022interoperable,curman2021automated,christou2022end,khan2020industrial}, these data were often unevenly distributed, noisy, lacking of labels, and contained a substantial amount of unstructured ones. As a result, these data were inadequate for training good deep learning models. Meanwhile, deep learning models were insufficient to handle large scale data with high efficiency.

Recent emergence of large scale foundation models\cite{bommasani2021opportunities,zhang2023comprehensive,zhao2023survey,ye2023mplug,sun2023aligning} were usually trained through extensive self-supervision and had exhibited robust generalization capabilities, exceptional zero-shot performance, and impressive multi-modal fusion abilities, as evidenced by successes in a wide range of downstream tasks, such as natural language processing, computer visions, etc.\cite{guo2023images,pan2023retrieving,harrison2023zero,liang2022expanding,chen2023position,ye2023mplugdoc,nag2022zero,liu2023zero}. Although the endeavor of using LSFMs to tackle challenges in intelligent manufacturing was just beginning, some progress had already been made attempted. \cite{moenck2023industrial,li2023chatgpt} discussed the potential applications of LSFMs in industrial manufacturing, but limited to specific industrial tasks or specific LSFM. Ji et al.\cite{ji2023sam} presented a quantitative comparison of the performance of vision foundational models in concealed scenarios with state-of-the-art deep learning models. Ogundare et al.\cite{ogundare2023resiliency} proposed a study on the resilience and efficiency of industrial automation and control systems generated by large language models (LLMs).

Although LSFMs showed great potentials in intelligent manufacturing, wherein powerful generalization ability, automatic high-quality training dataset generation and superior performance were highly desirable, studies in this areas remain in its early stages and a systematic review of LSFMs for intelligent manufacturing applications is still not available. This paper presented technical roadmaps for using LSFMs in intelligent manufacturing where deep learning methods met great obstacles. Our work aimed to provide guiding directions and discussions to help understand how LSFMs can benefit intelligent manufacturing.

The rest of this paper was organized as follows. Section \ref{DL challenges} described the challenges encountered by deep learning models in intelligent manufacturing. In Section \ref{Background fpr LSFMs}, we initially provided a brief overview of the current progresses of LSFMs, and subsequently we discussed the technical advantages of LSFMs in intelligent manufacturing that addressed the challenges faced by deep learning. Section \ref{Roadmaps and Cases} delineated the roadmaps of applying LSFMs in intelligent manufacturing. Finally, in Section \ref{Cases}, we illustrated how LSFMs could make progresses in intelligent manufacturing through several cases we applied in real manufacturing scenes.

%% file: Table/ML_DL_survey_list.tex
\begin{table*}[]
\begin{center}
\resizebox{0.8\linewidth}{!}{
\begin{tabular}{ccc}
\hline
Intelligent manufacturing areas         & Method           & List of survey papers \\ \hline
Prognostics and health management (PHM) & Machine learning & {\cite{gan2020prognostics,singh2020systematic,polverino2023machine,weikun2023physics,wang2017prognostics,tsui2015prognostics,souza2023machine,sutharssan2015prognostic,dogan2021machine,coulter2018intelligent,zhang2023top,fahle2020systematic}}               \\
                                        & Deep learning    & {\cite{rezaeianjouybari2020deep,zhang2019review,hu2022prognostics,ellefsen2019comprehensive,zhang2022prognostics,kumar2023prognostics,zhang2019review,remadna2018overview,thoppil2021deep,jayalaxmi2022machine,xu2022review,jamwal2022deep,wang2018deep,malhan2023role}}               \\ \hline
Quality control                         & Machine learning & {\cite{qin2022research,tran2022machine,dogan2021machine,zhang2023top,he2023research,fahle2020systematic,rahman2023machine,rajesh2022smart}}               \\
                                        & Deep learning    & {\cite{deng2023research,wu2022progress,zhao2021applications,di2023review,neupane2020bearing,ming2021application,xu2022review,jamwal2022deep,wang2018deep,malhan2023role}}               \\ \hline
Human action recognition              & Machine learning & {\cite{fahle2020systematic,scibilia2023modeling,ramasamy2018recent,qiu2022multi,nguyen2021trends,kulsoom2022review,attal2015physical,lara2012survey}}               \\
                                        & Deep learning    & {\cite{malhan2023role,chen2021deep,gu2021survey,zhang2022deep,nweke2018deep,ramanujam2021human,li2019survey,kumar2021human,khan2021survey}}               \\ \hline
Robotics                                & Machine learning & {\cite{fahle2020systematic,danthala2018robotic,liu2022robot,wang2019artificial,semeraro2023human,gulzow2020recent,sajwan2023review,domae2019recent}}               \\
                                        & Deep learning    & {\cite{xu2022review,aggarwal2022deep,soori2023artificial,de2021robotic,hernavs2018deep,valaskova2021deep,cordeiro2022bin,caldera2018review,duan2021robotics}}               \\ \hline
\end{tabular}
}
\caption{List of Survey Papers on Using Machine Learning and Deep Learning in Intelligent Manufacturing}
\label{tab:ML&DL_survey}
\end{center}
\end{table*}

%% file: Main_text/II_Previous_works_and_challenges_of_Deep_Learning_in_Intelligent_Manufacturing.tex
\section{Progresses and challenges of Deep Learning in Intelligent Manufacturing}
\label{DL challenges}

Pioneering work of artificial neural network started around Dartmouth workshops in 1956\cite{kline2010cybernetics}, and 1960s-1990s witnessed the development of machine learning when back propagation\cite{werbos1990backpropagation}, Boltzmann machine\cite{194417},K-nearest neighbor\cite{laaksonen1996classification},support vector machine\cite{vapnik1999overview}, among many other classical machine learning methods were invented to solve classification and regression issues in various scientific and industrial areas. 

The couple of decades around AD 2000 saw the evolution of deep learning which could extract non-linear and complicated features automatically from raw data with the helps of high-scale modeling and end-to-end optimization capabilities. Popular deep learning algorithms such as recurrent neural networks(RNN)\cite{hihi1995hierarchical}, long short-term memory(LSTM)\cite{memory2010long},convolution neural networks(CNN)\cite{lecun1998gradient}, deep belief network\cite{hinton2006reducing}, deep auto encoders, etc., were developed and widely used in various areas of intelligent manufacturing, e.g., prognostics and health management (PHM), quality control, human activity recognition, and industrial robotics.

\subsection{Progresses of deep learning in Intelligent Manufacturing}
\subsubsection{Prognostics and health management (PHM)}
PHM applies advanced technologies and methodologies for the monitoring and assessment of the health status of systems or equipments. The implementation of PHM aimed to avoid equipment failures and production halts, thereby enhancing production efficiency and reducing maintenance costs \cite{ochella2022artificial}. Traditional PHM methods primarily relied on machine learning technologies, including Support Vector Machines (SVM) \cite{tianle2010svm}, Random Forests \cite{wu2017data}, Principal Component Analysis (PCA) \cite{tayade2019remaining}, Particle Filtering \cite{jouin2016particle}, and Hidden Markov Models (HMM)\cite{soualhi2013prognosis}. These machine learning technologies often relied on expertise knowledge to manually select and extract desired features from data\cite{zio2013prognostics,lee2014prognostics}.

Nevertheless, with the manufacturing industry's progressive shift towards intelligent operations, novel challenges to traditional machine learning based PHM methods were introducted including data processing capacity, immediacy of response, and the adaptability to a broad range of scenarios.

In this context, deep learning methods, specifically those based on neural network models such as CNNs\cite{yang2020transfer}, RNNs\cite{vishnu2018recurrent}, and LSTMs\cite{wang2018remaining}, have demonstrated considerable improvements in PHM. For instance, Belmiloud et al. \cite{belmiloud2018deep} utilized Wavelet Packet Decomposition (WPD) alongside deep CNNs for feature extraction from bearing data, facilitating Remaining Useful Life (RUL) prediction. Li et al. \cite{li2019deep} and Zhu et al. \cite{zhu2018estimation} investigated multi-scale feature extraction techniques, wherein CNNs could represent various facets of the original data via interconnected convolutional and pooling layers more efficiently. Furthermore, some methods have enhanced PHM performance by combining different deep learning algorithms. For example, the CNN-LSTM model applied by Yue et al. \cite{yue2018end} was to address icing issues on wind turbine blades. Subsequently, Chen et al. \cite{chen2018learning} extended its application to the prognosis of the same issue, demonstrating its potential in resolving PHM challenges in specific industrial applications.

Although deep learning-based PHM strategies had achieved commendable progresses, they faced substantial challenges for broader applications \cite{tsui2015prognostics}, including the accurate acquisition and establishment of industrial datasets \cite{hagmeyer2021creation}, poor interpretability of PHM systems \cite{ishibashi2013gfrbs}, and lack of effective and stable capabilities to integrate with pre-existing industrial frameworks \cite{zhang2019review}. These identified challenges not only delineated the trajectory for future research but also underscored the need for more innovative solutions to propel the development and applications of PHM technologies.

\subsubsection{Quality control}
The core logics of traditional quality control methodologies was the utilization of statistical analysis and rule formulation to ensure adherence of product quality to established standards and requirements. In the early stages of development, Shewhart \cite{shewhart1929control} implemented control charts plotting sample means and standard deviations to monitor the state of control within production processes. The subsequent adoption of Six Sigma markedly reduced the rate of defects in industrial processes, establishing itself as a pivotal tool in traditional quality control \cite{joghee2017control}. However, these conventional methods harbored inherent limitations. Firstly their static nature rendered them ineffectual in adapting to changes and fluctuations within production processes, consequently impeding the transferability of quality control models \cite{ross2017total}. Additionally, their reliance on predefined rules and experiential knowledge proved insufficient in fully harnessing the extensive data generated in intelligent manufacturing, leading to inaccuracies in quality control and defect detection \cite{breyfogle2003implementing}. Furthermore, their capacity to process unstructured and image data was significantly constrained \cite{liu2018deep}.

In recent years, the focal point of research in the quality control domain has shifted towards the applications and enhancement of deep learning technologies. Deep learning models demonstrated exceptional capability in processing complex unstructured data, such as images and sounds, which significantly enhances their performance in product defect detection and classification \cite{bhatt2021image,tulbure2022review,wang2018deep}. Moreover, the profound feature learning ability of deep learning enables the automatic extraction of valuable features from extensive data sets, substantially improving the accuracy and efficiency of quality control measures \cite{elharrouss2022backbones,nikolados2022accuracy}. For instance, Zhang et al. \cite{zhang2020deep} developed an efficient method for surface defect detection in electronic components using Convolutional Neural Networks (CNN). In a similar vein, Essien et al. \cite{essien2020deep} employed LSTM models to model industrial production processes, achieving real-time quality control and fault prediction, thereby reducing downtime. Concurrently, Stricker et al. \cite{stricker2018reinforcement} proposed an adaptive control methodology based on deep reinforcement learning, capable of dynamically adjusting quality control strategies based on real-time data, underscoring the potential of deep learning in adaptive quality control applications.

In summary, deep learning has achieved significant milestones in the realm of quality control. It has not only enhanced the accuracy of product defect detection but also improved the efficiency and stability of production processes. These advancements have introduced novel tools and methodologies to the industrial sector, promising further improvements in product quality and reductions in production costs \cite{yang2020using,gorchet2020deep,lilhore2022deep}.

\subsubsection{Human action recognition(HAR)}
In intelligent manufacturing, HAR referred to the utilization of technology and methods to monitor, analyze, and identify the behaviors and movements of employees in work environment, and was applied to enhance safety, quality control, efficiency, and productivity. In recent years, deep learning technology had made significant strides in the field of HAR within industrial manufacturing\cite{31,32,33,34,35}. 
This progress was largely due to its potent feature learning capabilities and the ability to capture complex data patterns.
Manufacturers collected motion data from various sensors and utilized deep learning models such as CNNs\cite{park2022binary,38}, RNNs\cite{39}, and LSTMs\cite{shen2022human,38} to monitor and analyze workers' activities in real-time.
These HAR systems were capable of identifying unsafe work behaviors and issuing timely warnings, which reduced workplace injuries and increased the safety of the production environment.
For quality control, HAR helped in detecting and correcting operational errors by monitoring deviations from standard operating procedures, thereby reducing the incidence of defects.
Additionally, by analyzing workers' workflows, HARs could uncover inefficiencies within the production process, leading to the optimization of operational procedures and an increase in production efficiency.
In terms of equipment maintenance, HARs were monitored to prevent equipment failures and minimize downtime. HAR could also be used to identify training needs, providing a foundation for improving worker skills and minimizing operational errors.\cite{310,311,312,313}

On the technical front, multimodal learning emerged as a key development in HARs, enhancing the accuracy and robustness of activity recognition by combining visual and other sensor data. 
Transfer learning and domain adaptation techniques were employed to mitigate the annotation scarcity problem by leveraging pre-trained models and cross-domain knowledge\cite{abdulazeem2021human,an2021adaptnet}, thus reducing the dependency on extensive labeled data.
To address the class imbalance issue, researchers adopted strategies like resampling techniques, cost-sensitive learning, and ensemble learning. Unsupervised and semi-supervised learning methods also found applications in HARs, especially when labeled data was scarce\cite{an2021adaptnet,abdelbaky2021human}.
In summary, deep learning had shown enormous potential in the realm of industrial HAR. However, there was a continuous need to resolve challenges related to data, algorithms, and privacy\cite{314,315,316,317,318,319}.
Future research was expected to focus on developing more efficient algorithms, improving the generalization capabilities of models, and advancing the widespread applications of HAR technology in industrial production, meanwhile protecting worker privacy.

\subsubsection{Industrial Robotics}
Industrial robots were automated mechanical devices typically controlled by computer programs and used to perform various repetitive, precise, or high risk industrial tasks. These robots usually had multi-axis motion capabilities, equipped with various sensors and tools, capable of executing tasks in industrial fields such as manufacturing, assembly, and handling.
Deep learning (DL) technology had marked the advent of a new era of intelligence in manufacturing with its progress in industrial robotics applications. The core advantage of deep learning lay in its exceptional data processing and pattern recognition capabilities, which enabled robots to perform more complex and flexible tasks within intricate manufacturing settings\cite{41}\cite{42}.
For instance, by leveraging CNNs, robots had been capable of performing precise visual inspections, identifying manufacturing defects in real-time, and classifying them. This automated quality control not only improved product consistency but also significantly reduced the need for manual inspections.
In the realm of predictive maintenance, deep learning had enabled robots to anticipate potential equipment failures and carry out maintenance proactively by analyzing historical operational data. This was achieved through Recurrent Neural Networks (RNNs) and Long Short-Term Memory networks (LSTMs), which could handle and predict complex patterns in time-series data.\cite{43,44,45}.
Moreover, Deep Reinforcement Learning (DRL) technology was being utilized to develop autonomous robots capable of operating in unmanned or hazardous environments, thereby enhancing the safety and efficiency of operations.\cite{46}
Deep learning also played a crucial role in the control and optimization of assembly robots. Through deep learning algorithms, robots had been able to adapt to changing assembly conditions, collaborate with human operators, and learn from past experiences to improve future performance.
These technologies were also used to monitor the actions of robots, identifying potential safety risks and further enhancing workplace safety\cite{47,48}.

Despite the significant advantages that deep learning applications brought to the field of industrial robotics, such as increased production efficiency, waste reduction, improved product quality and safety, there were also challenges and drawbacks.
Deep learning models typically required large amounts of labeled data for training, which could be difficult and costly to obtain in real industrial environments. Additionally, the decision-making process of deep learning models often missed transparency, which could be problematic in industrial applications that demanded high reliability. Moreover, the computational intensity of deep learning algorithms could pose challenges for real-time applications\cite{49}\cite{410}.
In conclusion, the application of deep learning technology in the field of industrial robotics was achieving significant breakthroughs in the automation and intelligentization of production. However, to realize its full potential, further research and development were needed in the areas of algorithm interpretability, data efficiency, and computational optimization.

\subsection{Challenges of Deep Learning in Intelligent Manufacturing}

\input{Table/DL_challenges}

\subsubsection{Poor generalization capability}
\
\newline
\indent
\textbf{Overfitting.}
Overfitting refers to the phenomenon where a model excessively adapts to the training dataset during the training process, resulting in the model performing well on the training data but having poor generalization ability for new data. This issue should be particularly noted in the application scenarios of intelligent manufacturing because production manufacturing often involves relatively uniform samples and fixed tasks. Models trained under such conditions, even if they achieve high accuracy, should be suspected of being caused by overfitting. Overfitting severely damages the model's generalization, making even slight changes in samples or the environment likely to cause the model to fail.

The applications of deep learning algorithms in manufacturing encountered challenges, particularly in terms of poor generalization performance\cite{bubeck2023sparks,bhatt2021image,qi2019applying,weimer2016design,chen2020solar,mitash2023armbench} . The intricacies of manufacturing scenes introduced various domain shifts, often occurring in diverse and complex conditions. These shifts posed hurdles for deep learning applications, highlighting the need for robust adaptation strategies to ensure effective and reliable performance across the dynamic landscape of manufacturing processes.

\textbf{Poor domain knowledge integration.}
The challenge of poor domain knowledge integration stands as a significant hurdle in the application of deep learning to intelligent manufacturing. Domain knowledge, include expertise about specific manufacturing processes, materials, and industry-specific constraints, was invaluable for creating effective and dependable AI solutions. However, integrating this crucial knowledge into deep learning models was non-trivial for several reasons:

Implicit and tacit knowledge: Often, domain expertise resides within experts in a form that is implicit or tacit \cite{al2019multimodal,he2021deep,kilimci2019improved}, This posed a formidable challenge in terms of explicit articulation and encoding into deep learning models. This inherent complexity hindered the seamless integration of critical knowledge for effective deep learning applications in intelligent manufacturing.

Interdisciplinary nature of domain knowledge: Intelligent manufacturing often spanned multiple disciplines, such as mechanical engineering, materials science, automation, and more\cite{qi2019applying}. The integration of expertise from these diverse domains was inherently complex and necessitated a well-defined methodology for deep learning models. Effectively incorporating insights from various disciplines into the fabric of intelligent manufacturing posed a critical obstacle to achieve seamless and holistic domain knowledge integration.

Data-driven approach vs. Knowledge-driven approach: Deep learning models traditionally leaned heavily on training datasets to learn patterns, and finding a balance with the integration of crucial domain knowledge \cite{bhatt2021image} became challenging, especially in scenarios where labeled data were scarce. Striking an effective equilibrium between these contrasting approaches was essential for ensuring the robust integration of domain knowledge in intelligent manufacturing applications.

\textbf{Poor unstructured data representation.}   
Unstructured data encompassed information lacking of a predefined data model or specific organization. In manufacturing environments, this category included data from diverse sources such as images, videos, text documents, and sensor readings. The challenges of using poor unstructured data in the context of deep learning for intelligent manufacturing were discussed below. \par

Lack of standardization: The absence of standardization in data formats and labeling conventions increased the complexity of pre-processing and analysis when employing deep learning in intelligent manufacturing \cite{sun2018research}. The lack of uniformity across these aspects also posed a substantial obstacle to the seamless integration of unstructured data, requiring additional efforts in normalization and compatibility for effective utilization in deep learning models. \par

High dimensionality and high heterogeneity: The high dimensionality and high heterogeneity of data often existed in various levels of intelligent manufacturing of various time periods, products, fabricationlines, etc., such as images or videos, characterized by an extensive amount of information \cite{chen2021deep, yang2020using,pienaar2019human, he2021deep}, posed a significant challenge. Dealing with these complex datasets in the context of intelligent manufacturing required computationally intensive processes and extra-special processing. \par

Feature extraction: The task of feature extraction from unstructured data, involving the identification of meaningful patterns in images or relevant information from text, could be a non-trivial challenge demanding extra domain-specific expertise \cite{chen2021deep}. This required specialized knowledge to discern and capture pertinent features, emphasizing the complexity of representing unstructured data effectively for intelligent manufacturing. \par

Data volume and storage: The  volume of data generated by unstructured sources, such as images or videos, posed a significant logistical challenge in terms of efficient storage, retrieval, and processing \cite{khalil2021deep}. Dealing with these substantial datasets required careful considerations to ensure optimal management and utilization of deep learning.

\subsubsection{Limited high-quality training dataset}
\
\newline
\indent
\textbf{Limited labeling resources.} The challenge of limited high-quality training datasets for deep learning in intelligent manufacturing was accentuated by the scarcity of labeling resources. The process of annotating data for training could be exceptionally time-consuming with high costs, especially for special industrial dataset establishment wherein high expertise was needed. This scarcity in labeling resources imposed significant restrictions on both the size and diversity of the training dataset~\cite{manyar2023physics, manyar2022synthetic, jain2022synthetic, tabernik2020segmentation, he2019end, li2021defectnet, bhatt2021image, volkau2019detection, xu2020weakly}. 

\textbf{Variety and complexity.} The challenge of a limited high-quality training dataset could be further intensified by the variety and complexity nature of manufacturing environments, wherein diverse operating conditions, various product types, and a multitude of defect types might result in a high level of diversity in the training data~\cite{khalil2021deep}. Incorporating this diversity was essential for training deep learning models capable of effectively generalizing and adapting to the intricate and varied intelligent manufacturing scenarios   

\textbf{Imbalance and rarity.} The imbalance and rarity of certain faults or defects in the manufacturing process often existed wherein specific issues occurred infrequently resulting in severe class imbalance issues within the dataset, posing a significant challenge for deep learning models to deliver accurate performance~\cite{chen2021deep}.  \par

\textbf{Specificity and uniqueness.} Some manufacturing processes were highly specialized or even unique in particular industries\cite{nguyen2022enabling}. Acquiring relevant data for constructing a training set became particularly challenging when dealing with these processes. This specificity and uniqueness presented a barrier to assemble a comprehensive dataset, crucial for training deep learning models to effectively grasp the intricacies of these specialized manufacturing procedures.

\textbf{Anonymity and privacy concerns.} Anonymity and privacy concerns were significant for indusrtial companies. This restricted special algorithms like federal data sharing for research or other collaborative purposes, wherein this constraint significantly limited the availability of public datasets, crucial for efficient training of deep learning models~\cite{boulemtafes2020review}.

\textbf{Dynamic environments.} 
Dynamic environments in manufacturing ariseD as processes evolved over time, influenced by factors such as changes in raw materials, equipments and production techniques. This evolution introduced a mismatch between training datasets and data obtained from actual operating conditions~\cite{zonta2022predictive}. Adapting deep learning models to these dynamic shifts became difficult due to ever-changing manufacturing settings.

\subsubsection{Unsatisfactory performance}
\
\newline
\indent                  
\textbf{Single task, single modal} 
Most of the current deep learning algorithms could be effective only on a single task after sufficient training onspecially prepared datasets, as illustrated in Fig\ref{fig:overview}. When the same algorithm was applied on diffrtrnt tasks, using a different training dataset, its accuracy and efficiency usually decreased dramaticlly\cite{bhatt2021image,czimmermann2020visual,tabernik2020segmentation,li2021defectnet,volkau2019detection,xu2020weakly}. This restricted their performance in complex tasks and hindered their ability to effectively utilize diverse information. Even the deep learning models specifically designed for multitasking often only involved coupling multiple models together\cite{huang2020retracted,ouyang20193d}.

\textbf{Unsatisfactory accuracy} 
Accuracy referred to the ability of a model to make correct predictions or decisions. In manufacturing environments, accuracy was crucial for tasks such as defect detection, quality control, and process optimization. Although deep learning models achieved remarkably high accuracy on specific tasks in cases where there were sufficient high-quality training datasets\cite{zabin2023hybrid,choudhury2023adaptive}. Certain situations still significantly diminished the accuracy of deep learning, including insufficient data labeling, data fluctuations, excessively complex process objects, and noise interference, etc.\cite{he2019end,bhatt2021image,yang2020using,serradilla2022deep,aggarwal2022deep}

\textbf{Lack of interpretability} 
Interpretability referred to the ability to understand and explain the decisions made by a deep learning model. In many manufacturing scenarios, it was crucial to have a clear understanding of why a model made a particular prediction or decision, and a transparent decision-making process could facilitate subsequent improvements~\cite{chen2021deep} and aid in the identification and analysis of failures when they occurred~\cite{bhatt2021image}.

\textbf{Indusrtial network security}
In intelligent manufacturing, employed deep learning networks could be vulnerable to malicious adversarial attacks, wherein the utilization of adversarial samples, which where imperceptible to the human eye, might have the neural networks product entirely incorrect outputs~\cite{akhtar2018threat,zhang2020adversarial}. This posed challenges for the deployment of deep learning algorithms in practice, as such attacks might go unnoticed and cause direct financial losses to businesses~\cite{ibitoye2019analyzing,kumar2020adversarial}.

%% file: Table/DL_challenges.tex
\begin{table*}[]
\resizebox{1.0\linewidth}{!}{
\begin{tabular}{ccc}
\hline
\textbf{\begin{tabular}[c]{@{}c@{}}Challenges in the Application of Deep Learning in\\ \\ Intelligent Manufacturing\end{tabular}} & \textbf{Manifestation of challenges}                                & \textbf{References}                     \\ \hline
\multirow{3}{*}{Poor generalization ability}                                                                                      & Overfitting                                                         & {\cite{bubeck2023sparks,bhatt2021image,qi2019applying,weimer2016design,chen2020solar,mitash2023armbench}}   \\
                                                                                                                                  & Poor domain knowledge integration                                   & {\cite{al2019multimodal,he2021deep,kilimci2019improved,qi2019applying,du2017deeplog,bhatt2021image}}             \\
                                                                                                                                  & Poor unstructured data representation                               & {\cite{sun2018research,chen2021deep,yang2020using,pienaar2019human,he2021deep,chen2021deep,khalil2021deep}} \\ \hline
\multirow{6}{*}{\begin{tabular}[c]{@{}c@{}}Limited high-quality\\ \\ training dataset\end{tabular}}                               & Limited labeling Reources                                           & {\cite{manyar2023physics,manyar2022synthetic,jain2022synthetic,tabernik2020segmentation,he2019end,li2021defectnet,bhatt2021image,volkau2019detection,xu2020weakly}}                                \\
                                                                                                                                  & Variety and complexity                                              & {\cite{khalil2021deep}}                                \\
                                                                                                                                  & Imbalance and rarity                                                & {\cite{chen2021deep}}                                 \\
                                                                                                                                  & Specificity and uniqueness                                          & {\cite{nguyen2022enabling}}                                \\
                                                                                                                                  & Anonymity and  privacy concerns                                     &  {\cite{boulemtafes2020review}}                                       \\
                                                                                                                                  & Dynamic environments                                                & {\cite{zonta2022predictive}}                                \\ \hline
\multirow{2}{*}{Unsatisfactory performance}                                                                                       & Single task, single modal                                           & {\cite{bhatt2021image,czimmermann2020visual,tabernik2020segmentation,li2021defectnet,volkau2019detection,xu2020weakly,huang2020retracted,ouyang20193d}}                             \\
                                                                                                                                  & Unsatisfactory accuracy                                             & {\cite{he2019end,bhatt2021image,yang2020using,serradilla2022deep,aggarwal2022deep}}                \\
                                                                                                                                  & Lack of interpretability                                            & {\cite{samet2016primer,chen2021deep,bhatt2021image}}                 \\
                                                                                                                                  & Data security                                                       & {\cite{chen2021deep,akhtar2018threat,zhang2020adversarial,ibitoye2019analyzing,kumar2020adversarial}}                                  \\ \hline
\end{tabular}
}
\caption{Challenges in the Application of Deep Learning in Intelligent Manufacturing}
\label{tab:DL_challenges}
\end{table*}

%% file: Main_text/III_Background_for_LSFMs.tex
\section{Background for Large Scale Foundation Models}
\label{Background fpr LSFMs}

\subsection{Progresses on Large-Scale Foundation Models(LSFMs)}
Foundation models were designed to be trained on large-scale datasets, i.e., with parameters over billions to hundreds of billions and were named for the first time recently\cite{bommasani2021opportunities}. These models could have fixed most parameters after pre-training and adapted to a wide range of downstream applications through fine-tuning. In fact, large scale foundation models(LSFMs) had achieved revolutional progresses in the field of natural language processing\cite{openai2023gpt4}, computer vision\cite{kirillov2023segment} etc.

\begin{figure*}
    \centering
    \includegraphics[width=1\textwidth]{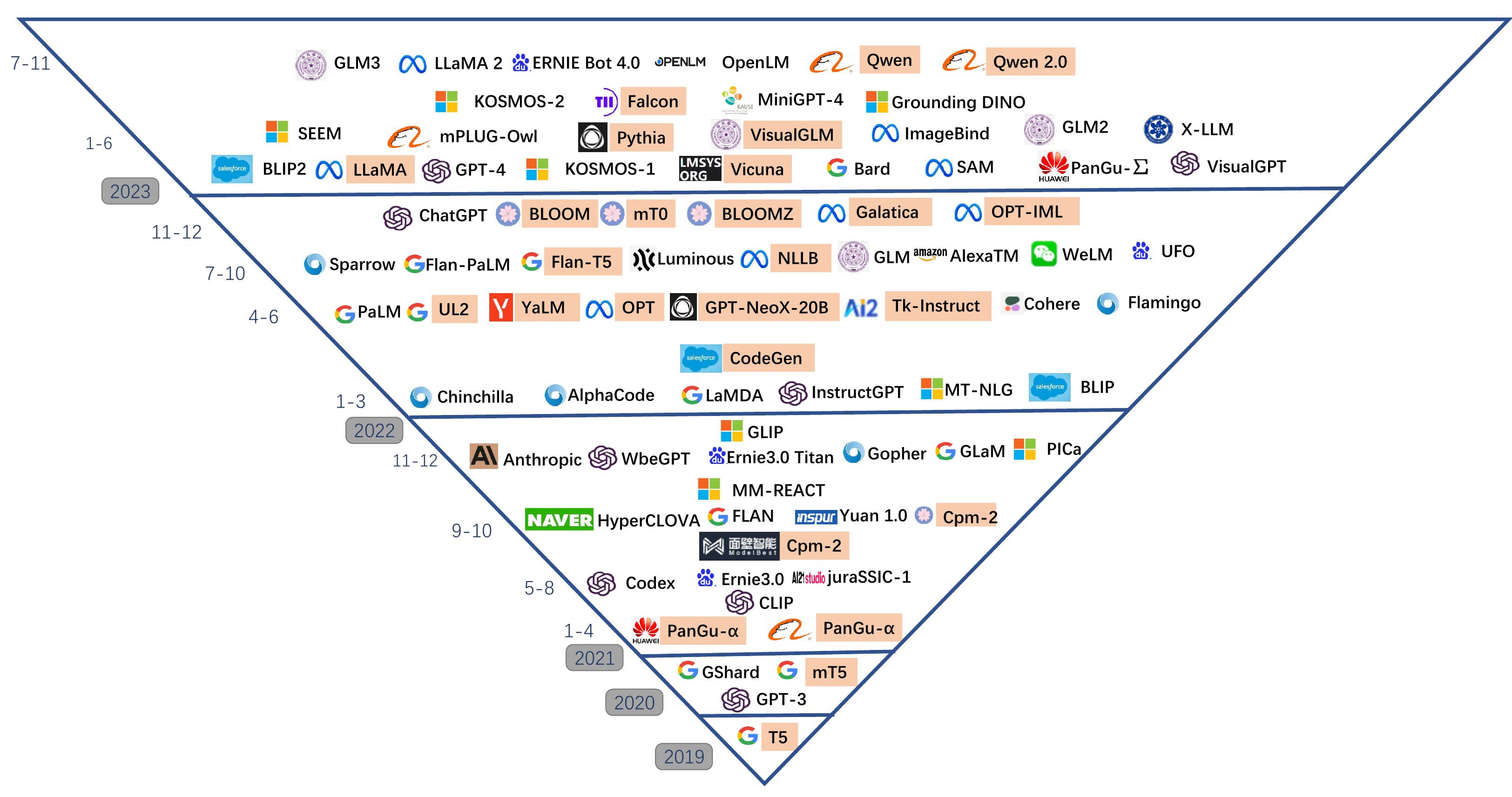}
    \caption{A timeline of existing large language models and large vision models}
    \captionsetup{}
    \label{fig:LLM timeline}
\end{figure*}

As depicted in Figure\ref{fig:LLM timeline}, the field of large language models (LLMs) has witnessed numerous impressive progressed\cite{brown2020language,chowdhery2022palm,taylor2022galactica,li2022competition}. Among these, the GPT series\cite{radford2018improving,radford2019language,brown2020language,openai2023gpt4} undoubtedly stood as the most renowned and benchmarked within LLMs. The latest version of the GPT series, GPT-4\cite{openai2023gpt4}, supported multimodal inputs, accepting both images and texts, and generated textual outputs. It is a Transformer-based model pretrained to predict the next token in documents. 

Subsequent fine-tuning processes could enhance its factual accuracy and ensure its performance to align with desired behaviors. In various professional and academic benchmark tests, GPT-4 had demonstrated performance comparable to human being levels, partcularly in fields such as human-machine interaction, education, healthcare, and law. The LlaMA model\cite{touvron2023llama} was currently the most popular open-source LLM, available in four sizes: 7B, 13B, 30B, and 65B. As LlaMA was pretrained on English corpora, it often necessitated fine-tuning with instructions or data from the target language when used, giving rise to a series of expanded models\cite{taori2023stanford,chiang2023vicuna,cui2023efficient} that constituted the LlaMA family.

For visual models, the Segment Anything Model (SAM)\cite{kirillov2023segment} had ushered computer vision into the era of LSFMs. SAM was composed of three main components: task, model, and data. Firstly, SAM's tasks involved designing a versatile model capable of transferring zero-shot capabilities to new image tasks and distributions. Inspired by ideas from LLMs, segmentation tasks were designed to be promptable so that it could refine valid segmentation masks from a given prompt. Secondly, SAM's model framework comprised three parts: an image encoder, a prompt encoder, and a mask decoder. The image encoder utilized a ViT\cite{dosovitskiy2020image} based on the MAE\cite{he2022masked}, wherein each image was processed once to obtain embeddings before running prompts. The prompt encoder handled sparse (points, boxes, text) and dense (masks) prompt inputs. The mask decoder employed a bidirectional Transformer decoder with self-attention and cross-attention mechanisms based on prompt images to map image embeddings, prompt embeddings, and output labelings onto masks. Lastly, recognizing the inadequacy of existing semantic segmentation datasets for training LSFMs, SAM introduced a data engine, a data generation approach involving iterative training-annotation cycles to obtain a substantial amount of annotated data. Establishing the SA-1B dataset, which consisted of 11 millions of images and 1.1 billions of masks with 99.1\% of them generated automatically, making it the largest segmentation dataset currently available. Compared to SAM, Segment Everything Everywhere All at Once(SEEM) model\cite{zou2023segment} offered broader interactional and semantic capabilities. It not only facilitated segmentation and category recognition in images and videos but also supported a wider range of input modalities, including points, bounding boxes, scribbles, text, speeches, and more. This attributed to SEEM's utilization of a unified prompting encoder, which encoded all types of prompts into a unified feature space. For the convenient deployment of visual foundation models, the unified feature optimization(UFO) model\cite{xi2022ufo} squeezed multiple tasks into a medium-sized model and further pruned it when transferring to down-stream tasks. This approach ensured flexible deployment of the UFO while retaining the advantages of large-scale pre-training.

Multimodal information processing was one of the important features and advantages of LSFMs and many multimodal methods were proposed recently for various downstream tasks. CLIP\cite{clip2021} connected text and images by creating a transferable visual model and taking a significant step towards breaking the traditional paradigm of computer vision. It learned the matching relationship between text-image pairs through contrastive pre-training methods, involving both Text Encoder and Image Encoder models. BLIP\cite{blip} and GLIP series\cite{li2022grounded,zhang2022glipv2}, made improvements based on CLIP by incorporating both image-text matching and image-text generation tasks during pre-training. This approach could generate images from text prompts or generate text descriptions from images. Grounding DINO\cite{liu2023grounding} combined the strengths of CLIP and DINO, enabling it to detect arbitrary objects based on human input. Diffusion model was also widely used in multimodal models\cite{nichol2021glide,saharia2022photorealistic,rombach2022high}. Inspired by non-equilibrium thermodynamics, the diffusion model applied forward diffusion to perturb the distribution of data and then learned to recover the data distribution through reverse diffusion.Not confined to aligning only images and text, DreamFusion\cite{dreamfusion} and Magic3D\cite{magic3d} linked text with 3D content, ImageBind\cite{imagebind} connected six data modalities: images, text, audio, depth, thermal data, and IMU data, forming a unified embedding space, and NExT-GPT\cite{wu2023nextgpt} could combine multimodal input to generate content. PaLM-E\cite{palme} combined language, vision, and robot commands for specific reasoning, visual language, and language tasks, texts could be generated to regulate low-level commands, facilitating control and planning of robots.

\subsection{Advantages of Large Scale Foundation Models(LSFMs)}
\subsubsection{Powerful generalization ability}
\
\newline
\indent
\textbf{Large scale datasets with high dimensions and complexity.} Although traditional deep learning models excelled in the applications of intelligent manufacturing, due to constraints in data scale and model expressive capabilities, these models could only perform well on specific tasks and were unable to give satisfactory performance on other tasks. In contrast, LSFMs, with significantly increased parameter quantities, e.g. ChatGPT with 1.42 trillions of parameters in its training dataset, greatly enhanced the model's expressive power, allowing for better modeling of general knowledge of massive training data containing enough variety, complexity, imbalance, rarity, specificity, uniqueness, anonymity, etc.. This meant that a single LSFM could adapt well to various downstream tasks, and researchers needed only to train one model, or even directly used publically available LSFMs to comprehensively address numerous industrial tasks. Bubeck et al.\cite{bubeck2023sparks} considered that, given the breadth and depth of GPT-4's capabilities, it could reasonably be regarded as an early version of Artificial General Intelligence (AGI) system, as it could tackle novel and challenging tasks spanning various fields such as mathematics, coding, visual, medical, legal, psychology, and demonstrated performance compareable to human being levels. Ji et al.\cite{ji2023segment} investigated the segmentation performance of SAM in anomaly surface defect detection tasks, and empirical evidence showed that SAM performed acceptably across different industrial tasks and human-machine interaction yielded even better results.

\textbf{Superior data human-machine interaction capability.} The emergence of multimodal LSFMs further enhanced the interaction capabilities of intelligent agents with humans, allowing for a more diverse range of tasks to be performed based on human prompts or instructions. SAM\cite{kirillov2023segment} was designed and trained to be promptable, supporting multimodal prompts such as text, keypoints, and bounding boxes, wherein users could easily segment specified objects using prompts like a point, a box, or a sentence. It even accepted input prompts from other systems, like visual focus information from AR/VR headsets, to select corresponding objects. The multimodal capability also allowed people to directly control robots using natural language\cite{palme,brohan2022rt,brohan2023rt,embodimentcollaboration2023open}, for example, by saying "Take the chips from the drawer," which involved multiple steps of action. In the field of robot operation, visual-language tasks, and language tasks, LSFMs trained for multitasking exhibited a higher level of performance compared to deep learning models trained for a single task. This led to high cross-task transfer efficiency in robot tasks, effectively extending the capabilities of robot control models and improving their performance and flexibility when handling complex tasks.

\subsubsection{Superior performance}
\
\newline
\indent
\textbf{Multi-task, multi-modal.} Compared to deep learning models designed for specific tasks on specific datasets, LSFMs had significant advantages and could integrate data from multi modals, achieving superior performance across multi tasks. Initially, the multi-task applicability of LSFMs significantly reduced research and development time and costs in practical intelligent manufacturing applications. Secondly, leveraging multi-modal inputs, which encompassed diverse data produced in manufacturing scenarios such as visual, linguistic, time series, and expert prior knowledge, often resulted in better performance compared to single-modal models\cite{gu2023anomalygpt,li2023myriad}. Moreover, the multi-task, multi-modal nature broadened the application boundaries of LSFMs in intelligent manufacturing. For instance, while traditional object detection algorithms could only detect fixed categories, LSFMs accomplished detection on open-set object\cite{liu2023grounding}, which undoubtedly better suited real-world application scenarios.

\textbf{Superior accuracy}. The improvements brought by LSFMs were evident in two aspects. On one hand, pre-trained models themselves already exhibited high accuracy. GPT-4, for example, achieved scores approximately in the top 10\% of test-takers in lawyer exams simulating\cite{openai2023gpt4}. Zero-shot CLIP outperformed fully supervised methods on 16 out of 27 tested datasets\cite{clip2021}. SAM delivered optimal performance in single-point effective masking and instance segmentation tasks, and its performance in object proposals surpassed baseline models for medium and large objects, as well as rare and common objects. It only fell short on small and frequent objects. In edge detection tasks, SAM lagged behind state-of-the-art edge detection algorithms but outperformed classical zero-shot transfer methods\cite{kirillov2023segment}.

On the other hand, while directly applying LSFMs in some scenarios might not be as effective as training dedicated neural networks for those specific application scenarios, LSFMs could further enhance their performance through fine-tuning. To address SAM's suboptimal performance when segmenting obscured, weakly bounded, low-contrast, small, and irregularly shaped target objects, Wang et al.\cite{wang2023detect} used shadow data combined with sparse cues for fine-tuning SAM. They also incorporated a Long Short-Term Memory network, surpassing state-of-the-art techniques\cite{liu2023scotch} and improving the Mean Absolute Error (MAE) and Intersection over Union (IoU) by 17.2\% and 3.3\%, respectively. For medical images typically characterized by weak boundaries and irregular shapes, SAMed\cite{zhang2023customized} applied a low-rank-based (LoRA) fine-tuning strategy to SAM's image encoder and finetuned it together with the prompt encoder and the mask decoder on labeled medical image segmentation datasets. It performed on par with state-of-the-art methods\cite{azad2023dae,oktay2018attention} on the Synapse multi-organ segmentation dataset.

\textbf{Logical reasoning ability.} For LSFMs, it was discovered that when model size exceeded a certain threshold, the model could offer unpredictable emergent abilities, such as autonomous logical reasoning abilities. This was referred to as Chain-of-Thought (CoT)\cite{kojima2022large,suzgun2022challenging}, which wasn't simply about constructing input-output pairs as prompts but involved incorporating intermediate reasoning steps into the prompts, leading to the final output. In essence, LSFMs would break down complex tasks into finer sub-tasks and then formulate strategies to accomplish these sub-tasks. Leveraging this feature, LSFMs could tackle more complex problems that were previously challenging for deep learning models. Suzgun et al.\cite{suzgun2022challenging} selected 23 challenging tasks from the BIG-Bench evaluation method\cite{srivastava2022beyond}, including causal reasoning, word sorting, and action comprehension, wherein prior language model evaluations did not surpass average human performance. However, when CoT was applied, Pa-LM\cite{palme} exceeded human average performance in 10 tasks, and Codex\cite{chen2021evaluating} surpassed human average performance in 17 tasks. In reality, LSFMs could assist humans in performing more advanced activities\cite{choudhary2023complex}, such as tasks related to intelligent manufacturing, like production planning and material allocation scheduling by employing logical reasoning ability of LSFMs.

\textbf{Superior analysis capability .} Natural language is one of the most common and frequent means of information exchange among humans. The powerful natural language capabilities of LLMs implies many potential application advantages. Unlike explicit instructions, natural language can be ambiguous and carry multi-layered information. \cite{bubeck2023sparks} discovered that LSFMs not only had the ability to understand natural language but also could infer the underlying beliefs, emotions, desires, intentions, and knowledge, among other psychological states behind the language. As a crucial aspect of intelligence, in addition to understanding others' language, LSFMs also had the capability to comprehend their own behavior to some extent. In \cite{bubeck2023sparks}, the quality of explanations was assessed using output consistency and process consistency, wherein GPT-4 exhibited reliable output consistency with a lack of processing consistency, providing reasonable explanations about how predictions were made and thus deepening the understanding of the tasks themselves.

\subsubsection{Automatic high-quality training dataset generation}
\
\newline
\indent
\textbf{Semi-supervised / self-supervised}. Self-supervised methods had already been used in various LSFMs for automatic training dataset establishment. The self-supervised training mode of LSFMs significantly reduced the need for accurate data labels, making it easier to obtain large amounts of unlabeled data. For instance, BERT\cite{devlin2019bert} employed portions of unannotated text from the BooksCorpus and Wikipedia. It utilized two self-supervised tasks for training, and the trained model achieved optimal performance on 11 downstream tasks through fine-tuning. CLIP\cite{clip2021} used 400 million "image-text pairs," while Wu et al.'s Text-to-Image 2.0\cite{fei2022artificial} employed 650 million "image-text pairs" for training. Additionally, the automatically labeled data generated during self-supervised training could be collected for wide research and industrial applications. For example, the SAM model\cite{kirillov2023segment} built the largest segmentation dataset to date, known as SA-1B which had 1.1 billions of masks and 99.1\%of them was geb=nerated automatically. In fact, it became a classic dataset for future computer vision segmentation model training and evaluation, whereinover 300academic papers used SAM and SA-1B for various segmentation tasks after its release in March 2023.

\textbf{Superior generative capability.} The generative LSFMs\cite{rombach2022high,openai2023gpt4} changed the landscape of AI by generating new data through learning the distribution of existing data, giving AI a sense of "creativity." Generative LSFMs were not only used for generating dialogues and images but also found applications in smart manufacturing to automatically generate high-quality training data. Whitehouse et al.\cite{whitehouse2023llmpowered} attempted to expand multilingual commonsense reasoning datasets using LLMs, integrating datasets generated by LLMs significantly enhanced the performance of the trained model. Sen et al.\cite{sen2023people} utilized LLM to generate Counterfactually Augmented Data (CADs), which was commonly employed to train models for robustness against false features, achieving performance almost comparable to manually generated CADs.

\input{Table/LSFM_advantages}

%% file: Table/LSFM_advantages.tex
\begin{table*}[]
\resizebox{1.0\linewidth}{!}{
\begin{tabular}{cccc}
\hline
\begin{tabular}[c]{@{}c@{}}The advantages of using LSFM \\ in intelligent manufacturing\end{tabular}          & Characteristics of LSFM                                                                                              & Reference                  & Description                                                                                                                                                                                                                                                       \\ \hline
\multirow{7}{*}{Powerful generalization ability}                                                              & \multirow{4}{*}{\begin{tabular}[c]{@{}c@{}}large-scale datasets with \\ high dimensions and complexity\end{tabular}} & {\cite{bubeck2023sparks}}                   & \begin{tabular}[c]{@{}c@{}}Extensive testing of GPT's abilities in tasks \\ such as artistic creation, coding, mathematics, and embodied intelligence has been conducted, \\ and it is believed that GPT-4 can be considered an early version of AGI\end{tabular} \\
                                                                                                              &                                                                                                                      & {\cite{bommarito2023gpt}}                   & Experimentally evaluate GPT on both sample Regulation (REG) exam and the Unified CPA Exam                                                                                                                                                                         \\
                                                                                                              &                                                                                                                      & {\cite{ji2023segment}}                   & \begin{tabular}[c]{@{}c@{}}Examined SAM's performance across various fields, \\ acknowledging its strong generalization \\ while emphasizing the pivotal role of quality prompts for accurate segmentation.\end{tabular}                                          \\
                                                                                                              &                                                                                                                      & {\cite{feng2023cheap}}                   & \begin{tabular}[c]{@{}c@{}}Using an example to guide the synthesis module and LoRA fine-tuning strategy, \\ effectively fine-tuning SAM in the medical field using only a small amount of labeled data\end{tabular}                                               \\
                                                                                                              & \multirow{3}{*}{superior human-machine interaction capability}                                                             & {\cite{kirillov2023segment}}                   & \begin{tabular}[c]{@{}c@{}}Users can use methods such as points, boxes, or sentences to specify segmentation objects, \\ and even use input prompts from other systems\end{tabular}                                                                               \\
                                                                                                              &                                                                                                                      & {\cite{palme,brohan2022rt,brohan2023rt,embodimentcollaboration2023open}} & \begin{tabular}[c]{@{}c@{}}Using natural language to directly control robots to complete tasks \\ involving multiple action steps\end{tabular}                                                                                                                    \\ \hline
\multirow{10}{*}{Superior performance}                                                                        & \multirow{2}{*}{\begin{tabular}[c]{@{}c@{}}multi-task,\\ multi-modal\end{tabular}}                                   & {\cite{gu2023anomalygpt}}                  & \begin{tabular}[c]{@{}c@{}}A conversational Industrial anomaly detection VLM,\\  which can achieve high accuracy recognition of a wide range of anomaly types \\ on a wide range of detected objects with just one shot\end{tabular}                              \\
                                                                                                              &                                                                                                                      & {\cite{li2023myriad}}                    & \begin{tabular}[c]{@{}c@{}}Integrating multiple modalities such as vision, language, and expert prior knowledge, \\ it exhibits high performance in a wide range of detection tasks\end{tabular}                                                                  \\
                                                                                                              & \multirow{4}{*}{superior accuracy}                                                                                   & {\cite{clip2021}}                   & Zero shot CLIP performs better than fully supervised methods on 16 out of 27 test datasets                                                                                                                                                                        \\
                                                                                                              &                                                                                                                      & {\cite{liu2023deidgpt}}                   & \begin{tabular}[c]{@{}c@{}}GPT for zero-shot recognition of medical texts,\\ showed the highest accuracy and remarkable reliability\end{tabular}                                                                                                                  \\
                                                                                                              &                                                                                                                      & {\cite{wang2023detect}}                   & \begin{tabular}[c]{@{}c@{}}Fine-tuning SAM by combining shadow data with sparse cues, coupled with LSTM, \\ surpassed the previous state-of-the-art methods\end{tabular}                                                                                          \\
                                                                                                              &                                                                                                                      & {\cite{zhang2023customized}}                   & \begin{tabular}[c]{@{}c@{}}Fine-tuning SAM's image encoder on a medical image segmentation dataset using LoRA \\ achieved performance comparable to state-of-the-art methods.\end{tabular}                                                                        \\
                                                                                                              & \multirow{2}{*}{logical reasoning ability}                                                                           & {\cite{choudhary2023complex}}                   & \begin{tabular}[c]{@{}c@{}}Use LLM formulates complex KG reasoning \\ as a combination of contextual KG search and logical query reasoning\end{tabular}                                                                                                           \\
                                                                                                              &                                                                                                                      & {\cite{suzgun2022challenging}}                   & \begin{tabular}[c]{@{}c@{}}23 challenging tasks were selected from the BIG Bench evaluation method, \\ and after employing CoT, LSFMs surpassed human average performance in half of these tasks.\end{tabular}                                                    \\
                                                                                                              & \multirow{2}{*}{superior analytical capability}                                                                        & {\cite{bubeck2023sparks}}                   & \begin{tabular}[c]{@{}c@{}}LSFM not only understands natural language, \\ but also infers the underlying psychological states behind language\end{tabular}                                                                                                        \\
                                                                                                              &                                                                                                                      & {\cite{singh2023explainingdata}}                  & \begin{tabular}[c]{@{}c@{}}Using LLM to discover and interpret patterns in data, \\ generating interpretable automatic prompts\end{tabular}                                                                                                                       \\ \hline
\multirow{5}{*}{\begin{tabular}[c]{@{}c@{}}Automatic high-quality\\ training dataset generation\end{tabular}} & \multirow{3}{*}{semi-supervised / self-supervised}                                                                   & {\cite{kirillov2023segment}}                   & Utilize a data engine to construct the largest segmentation dataset to date (SA-1B)                                                                                                                                                                               \\
                                                                                                              &                                                                                                                      & {\cite{clip2021}}                   & Training using 400 million weakly supervised 'image text pairs' obtained from the internet                                                                                                                                                                        \\
                                                                                                              &                                                                                                                      & {\cite{fei2022artificial}}                   & \begin{tabular}[c]{@{}c@{}}Utilize 650 million weakly semantically related images and texts on the internet \\ to train data without requiring any manual annotation\end{tabular}                                                                                 \\
                                                                                                              & \multirow{2}{*}{superior generative capability}                                                                      & {\cite{sen2023people}}                    & Verified the effectiveness of LLM generating Counterfactually Augmented Data (CADs)                                                                                                                                                                               \\
                                                                                                              &                                                                                                                      & {\cite{whitehouse2023llmpowered}}                    & Augment the multilingual commonsense reasoning datasets using LLM's generation                                                                                                                                                                                    \\ \hline
\end{tabular}
}
\caption{The advantages of using LSFMs in intelligent manufacturing}
\label{tab:LSFM_advantages}
\end{table*}

%% file: Main_text/IV_Roadmaps_for_solving_problems_with_LSFMs.tex
\section{Roadmaps of LSFMs for intelligent manufacturing applications}
\label{Roadmaps and Cases}

\subsection{Roadmaps for powerful generalization ability}
\subsubsection{Pre-training combined with fine-tuning}
As model parameter and size surpassed a certain threshold, these models not only exhibited emergent performance improvements but also acquired functionalities like logical reasoning abilities, absent in smaller-scale models~\cite{brown2020language,han2022survey}. The combination of pre-training with fine-tuning in LSFMs offered various possibilities to address issues encountered by traditional small-scale deep learning methods in intelligent manufacturing.

LSFMs pre-trained on diverse general datasets reduced their reliance on limited, task-specific datasets, hence mitigating overfitting risks despite the large parameter count of the models. Kahatapitiya et al.~\cite{kahatapitiya2023victr}, acknowledging the limited availability of video-text matching data, applied a pre-trained image-text model to the video domain for video-text matching instead of training from scratch. Additionally, specific fine-tuning strategies could enhance model generalization to further avoid potential model overfitting during fine-tuning with small sample learning. Song et al.~\cite{song2023fdalign} proposed a fine-tuning method called Feature Discriminant Alignment (FD Align) to enhance model generalization by maintaining consistency in pseudo-features, demonstrating effectiveness in within-distribution (ID) and out-of-distribution (OOD) tasks.

\subsubsection{Building structured data through LSFMs}
LSFMs cloud be used to extract and comprehend intricate unstructured data, encoding it into manageable structured formats, for instance, handling unstructured text data within work orders~\cite{sobhkhiz2023integrating}. Deep Generative Models (DGM) and models like VIT~\cite{dosovitskiy2020image} were designed to uncover complex high-dimensional probability distributions from unstructured data to extract more abstract, more intricate features. Oliveira et al.~\cite{oliveira2023new} outlined four types of DGM: Energy-based Models (EBM), Generative Adversarial Networks (GAN), Variational Autoencoders (VAE), and Autoregressive models, and how they were applied to SCM optimization.

\subsubsection{Embedding knowledge through prompts}
Once expert knowledge was encoded, it could be fused with input texts or image features, thereby improving the accuracy of output~\cite{wang2021knowledge}. Many LSMFs, such as ChatGPT and SAM, inherently included manual prompt encoding, allowing for the fusion of domain knowledge without modifying the models through prompts. For example, for abstract human behavior activities, it might be difficult for models to describe them all at once. Therefore, it could be guided to generate activity descriptions related to objects initially, emphasizing crucial objects to distinguish similar activities. Subsequently, it could identify the activity classes of human activities and help explain the contexts~\cite{xia2023unsupervised}. Furthermore, LSFMs could even gather relevant domain knowledge during training by collecting case studies~\cite{zhao2023expel}.

\subsubsection{Using multimodal LSFMs}
Intelligent manufacturing, usually yielded a multiple forms of data, encompassing free-text maintenance logs, images, audio and video recordings. The inherent diversity of these data posed a formidable challenge for singular modalities within deep learning models. LSFMs such as Visual-GPT~\cite{chen2022visualgpt} and ImageBind~\cite{imagebind} had emerged as viable solutions. These models exceled in the simultaneous encoding of a spectrum of data, including images, text, audio, depth, thermal, IMU data, and time-series signal data~\cite{li2023multimodal,garza2023timegpt}. This expanded capacity could not only enrich the range of data captured in intelligent manufacturing but also endow LSFMs with distinct functionalities such as cross-modal retrieval, modal fusion via arithmetic operations, and cross-modal detection and generation. Leveraging these expansive LSFMs facilitates the precise handling of unstructured data and the synthesis of diverse structured data sources. In complex industrial environments characterized by multiple disturbances, LSFMs demonstrated enhanced robustness compared to conventional single-mode deep learning methods.

\subsubsection{Regularization and ensemble learning}
LSFMs can solve overfitting problems through methods such as regularization, and ensemble learning. Regularization can limit the complexity of the models, pruning can remove unnecessary nodes and connections, and ensemble learning can combine the prediction results of multiple models to improve the generalization ability of the models.

Although many LSFMs like GPT-3 and PaLM did not use dropout~\cite{srivastava2014dropout} during training, it still had a significant effect on LSFMs. For instance, by using dropout during the training process Galactica~\cite{taylor2022galactica} achieved a 120 billion parameter model without overfitting. Moreover, to alleviated the reduction in training speed for LSFMs due to dropout, gradually introducing dropout into the process during training could yield performance comparable to using dropout consistently throughout~\cite{xue2023repeat}.

\subsubsection{Continual Learning/Life-Long Learning}
Most current deep learning models in intelligent manufacturing assumed that the normal mode remained unchanged. However, variations in the manufacturing environment occurred frequently.Continuous learning/lifelong learning involves acquiring and identifying new knowledge while retaining previously learned knowledge. LSFMs possessed robust capabilities for continuous learning by collecting past task outcomes as experience. 
Through this process, LSFMs continually enhanced themselves using prior knowledge~\cite{wang2022learning,zhao2023expel}. The continual learning feature of LSFMs allowed them to continually accumulate new knowledge during actual production processes to adapt to potential changes in complex real-world environments\cite{wang2022learning,zhao2023expel}. This ability was beneficial in preventing models trained on fixed patterns from experiencing overfitting. Applying specific constraints to this process further enhances the model's performance and stability~\cite{yadav2023exploring}.

\subsubsection{LSFM assisted construction of knowledge graph}
Knowledge graphs are forms of expression that acquires knowledge by understanding graph structures~\cite{fensel2020introduction}, and were highly effective. However, knowledge Graph Engineering (KGE) required an in-depth understanding of graph structure, logic, and knowledge content, causing a lot of work. The contextual comprehension and representation capabilities of deep learning methods were unsatisfied, especially when encountering entirely new or rare knowledge. Using the knowledge comprehension abilities and advanced reasoning skills of LLMs, it was possible to automatically generate knowledge graphs in professional domains\cite{meyer2023llm}, and was expected to enhance the model's understanding of specific domain knowledge achieved by synergizing the knowledge graphs with pre-trained language models\cite{chen2020review}.

\subsection{Roadmaps foor automatic high-quality training dataset generation}
\subsubsection{Generation of higher-quality datasets}
Generative models such as diffusion could potentially facilitate the generation of higher-quality synthetic data compared to traditional data synthesis methods~\cite{rombach2022high}. Using a text-to-image diffusion model could generated realistic image variations for data augmentation. Unlike simple augmentation methods such as concatenation, rotation, flipping, augmentation based on the diffusion model could alter higher-level semantic attributes, like the paint job on a truck\cite{trabucco2023effective}. To addressed the issue of requiring a large amount of data for training the diffusion model itself, Wang and colleagues transformed the two-dimensional diffusion model into three dimensions using chain rules, enabling the generation of three-dimensional object data\cite{wang2023score}. Moreover, Transform could be used to weighted average or score the results of multiple prediction models, and to learn and simulate historical data to obtain more robust prediction results.

In Section V, we demonstrated how we used LSFMs to achieve low-cost, automated action recognition data annotation on industrial production lines. 

\subsubsection{Improving data quality}
High-quality data was crucial for both model training and decision-making in intelligent manufacturing, wherein raw data often had issues such as missing values, outliers, and repeated values. LSFMs could be used to automatically remove impurity data, reduce prediction errors, and improve data quality. For instance, BLIP \cite{blip} relied on intermediate training models to automatically remove poorly matched image-text pairs from the dataset during training and improved the text annotations of certain images. Jain et al.\cite{jain2023llmassisted} using LLM to enhance the structural organization and readability of the code used as training data. Models fine-tuned on the cleaned data perform significant improvements in the algorithmic code generation tasks.

\subsubsection{Zero-shot and few-shot}
A major challenge in industrial defect detection is the lack of abnormal samples, and the abnormal situations of industrial products are often diverse and unpredictable. The LSFM could effectively achieve zero sample detection or few sample detection. Gu et al.~\cite{gu2023anomalygpt} explored the use of Large Vision-Language Models(LVLMs) to solve industrial anomaly detection problems and proposed a new LVLM based industrial anomaly detection method, AnomalyGPT. On the MVTec anomaly detection dataset, AnomalyGPT could achieve the most advanced performance of 86.1\% accuracy, 94.1\% image level AUC and 95.3\% pixel level AUC with only one normal shot. This application method no longer requires to collect abnormal samples or create datasets for each task to train specific models, and only needed a small amount of data fine-tuning to achieve good detection results. In predictive maintenance, for instance, Leite et al.~\cite{leite2023detecting} employed LLMs to classify credibility signals, which were commonly used to assess the authenticity of predictive contents. The LLM-based approach outperformed state-of-the-art classifiers on two misinformation datasets without the need for any ground-truth labels.

\subsubsection{Pre-training combined with fine-tuning}
Altough some preliminary work provided datasets for intelligent manufacturing scenarios such as HAR~\cite{garcia2020public,liu2017pku,shahroudy2016ntu}, quality control\cite{huang2019pcb,feng2021x,lv2020deep}, and PHM~\cite{lessmeier2016condition,center2018case}, these datasets were characterized by small scales, narrow coverage, singular scenes, simple operating conditions, and uneven data distributions. LSFMs pre-trained on extensive data could identify general features of real-world entities, providing an efficient solution for achieving accurate and flexible intelligent manufacturing in data-limited environments~\cite{brown2020language}. Pre-trained models trained on large scale data were then fine-tuned on smaller-scale data to enhance model accuracy and generalizability. For instance, Sun et al.~\cite{gao2023desam}. used BERT in medical texts and achieved good performance with just a small dataset for fine-tuning. Similarly, Radford et al.~\cite{radford2019language}. demonstrated GPT's transfer learning abilities across various tasks.

\subsection{Roadmaps for superior performance}
\subsubsection{Improvements by prompts}
Generally, after training was completed, deep learning models no longer accepted 'guidance' and instead operated on the basis of trained parameters for inference. However, LSFMs had superior data integration capability that could enhance output performance by leveraging various forms of prompts. Ji et al.~\cite{ji2023segment} found that the quality of prompts had a crucial impact on the accuracy of LSFMs. To address SAM's suboptimal segmentation performance on small scales and irregular boundaries, multiple prompts could be employed to derive more precise segmentation outcomes from distributions\cite{zhang2023segment}. Specifically, Deng et al.~\cite{deng2023samu} utilized Monte Carlo simulations with prior distribution parameters to estimate SAM's predicted distribution. This approach allowed for estimating arbitrary uncertainties by considering multiple predictions from individual images. Alternatively, networks could also be employed to acquire enhanced cues, generating augmented cues from inputting original cues to yield masks and subsequently outputting augmented cues. By merging these cues as new hints, segmentation performance could be enhanced~\cite{dai2023samaug}. It is also worth carefully handling decoupling mask generation and prompt embedding to prevent misleading prompt from having adverse effects on mask generation~\cite{gao2023desam}.

\subsubsection{Enhanced input data}
In LSFMs, the term "foundation" indicating that LSFMs could easily be used as the foundation to combine with other algorithms. This ensured that even in situations where LSFMs performed unsatisfactorily when being used alone, good performance could still be guaranteed by combining them with other algorithms. VLM exhibited strong robustness against various corruptions, yet some corruptions led to model performance degradation, such as corruption related to blur~\cite{huang2023robustness}. Additionally, SAM's performance was proven to be inadequate in concealed and camouflaged scenes~\cite{ji2023sam,tang2023can}. Fortunately, extensive prior research has been conducted in both deblurring~\cite{sun2015learning,aslam2022removal,huang2019image} and target detection techniques for concealed and camouflaged scenes~\cite{fan2021concealed,he2023camouflaged}. As one of the characteristics and advantages of LSFMs, VLM could effortlessly combine with other models, using preprocessed data as input or incorporating detection boxes from other object detectors as prompts.

\subsubsection{Cross modal pre-training}
LSMFs overcome the limitations of single task and single modal in deep learning, and could achieve multi-task and multi-modal applications with unified model, after cross modal pre-training\cite{gu2022wukong}. By leveraging contrastive loss during training to establish correlations between image and text features, it was possible to achieve open-set object recognition and detection~\cite{li2022grounded,zhang2022glipv2,liu2023grounding}. This could prevent tasks from being constrained by pre-defined categories in training. To achieve satisfactory pre-training performance, success relied on both the scale of cross-modal datasets~\cite{gu2022wukong,xie2023ccmb} and the model's ability to leverage weakly aligned data~\cite{zhou2023rc3}. Li et al.~\cite{li2023enhanced} employed a pre-trained model for weakly supervised label classification to measure semantic similarity in videos within an industrial system. By incorporating an enhanced cross-modal Transformer block, they maximized the utilization of interactive information between video and texture features.

\subsubsection{Pre-training combined with fine-tuning}
Compared to the unsatisfactory accuracy achieved by deep learning in situations with limited data and complex processes, large scale pre-training not only bestows powerful generalization capabilities on LSFMs but also endows them with the potential for higher accuracy~\cite{brown2020language,han2022survey}. While directly using pre-trained LSFMs may not always outperform specially designed deep neural networks~\cite{ji2023segment}, fine-tuning them with specific intelligent manufacturing domain dataset data effectively can improve their accuracy~\cite{hu2023skinsam,chu2023finetune}, potentially surpassing existing deep learning models. Technologies like P-Tuning~\cite{DBLP:journals/corr/abs-2110-07602}, Lora~\cite{hu2021lora}, QLora~\cite{dettmers2023qlora}, facilitated the fine-tuning process in LSFMs.

Apart from that, training deep learning models on integrated datasets increased the privacy risk of data leakage. Using pre-trained LSFMs emerged as a viable solution to enhance data security, reducing the privacy risk brought by the extensive data requirements for training models from scratch. These pre-trained models could achieve effective results with minimal fine-tuning, thereby reducing exposure to sensitive data. During the fine-tuning stage, a limited segment of the LSFM network required adjustment, introducing differential privacy techniques. Specifically, techniques proposed by Abadi et al.~\cite{abadi2016deep} were applied during the fine-tuning process. These measures could maintain the privacy of the data involved in fine-tuning LSFM, ensuring a more secure training environment.

\subsubsection{Adopting distributed learning}
Data in manufacturing are not as readily available as those in natural language and other areas, hence the adoption of distributed learning methods~\cite{mcmahan2017communication} could be beneficial in both training and security aspects for LSFMs used in intelligent manufacturing wherein data for training could be obtained from different lines, factories, or even countries. Distributed learning methods, such as federated learning, involved the local processing of data from each party, with only intermediate results (like gradients) being aggregated for model updates. This enabled clients (devices or organizations) to collaboratively train machine learning models without exposing their data, greatly increasing data usage efficiency~\cite{fan2023fatellm,kuang2023federatedscope}. Combining with such technologies could empower LSFMs to surpass traditional methods not only in performance but also in providing a more secure data processing framework when dealing with sensitive indusrtial information.

\subsubsection{Using LSFM's own output for interpretation}
Deep learning models were often considered as 'black boxes' due to their decision-making processes being highly abstract and non-intuitive. LSFMs, especially LLMs, demonstrated remarkable contextual comprehension in tasks, hence attempting to use LLMs to explain models was a potentially feasible. In a study by Bubeck et al.~\cite{bubeck2023sparks}, it was found that LLMs exhibited strong result consistency in their outputs, implying that the model followed a fixed “thinking” pattern. Therefore, asking questions like “Please explain the reasons behind your prediction” to chat-GPT was proven to be effective, when being preceded by reasonable prior questions.
This idea could also be applied to models based on encoder structures~\cite{MaskedAutoencoders2021}, to address the limitations of AE by performing bias analysis on reconstructed input features to obtain explanations~\cite{9658550}.

\subsubsection{Using LLM to explain other models}
LLMs had powerful text capabilities, and it was envisioned to leverage the knowledge obtained from LLMs to interpret other neural networks. To achieve this, LLMs were used to summarize and score the outputs of the models under analysis~\cite{singh2023explaining}. In addition, LLMs could be employed to generate or match counterfactuals, simulating or estimating different choices in an event or behavior to better understand the model's predicted outcomes~\cite{gat2023faithful}. Alternatively, embedding LLMs directly into the model training could allow for efficient inference while achieving good interpretability~\cite{singh2022augmenting}.

\subsubsection{Visualize the running process}
Enabling the extraction of intermediate feature maps from neural network outputs could assist in understanding the features being focused on, even though these feature maps could be still highly abstract. Visualizing attention could provide a more intuitive explanation over feature maps by employing self-attention mechanisms with token-linking in its architecture. The strength of the attention links could intuitively be considered as indicators of each token’s contribution to classification. The visualizing attentions aided in understanding the interested parts to the models~\cite{chefer2021transformer}. Considering that LSFMs are mostly built on transformer structures, visualizing the output of attentions to improve the interpretability of LSFMs is promising.

Moreover, it was possible to visualize the data through the output of LSFMs with the helps of interpretable auto prompting~\cite{singh2023explainingdata}. For intelligent manufacturing LSFMs could integrate scattered and complex supply chain data and present the data through data analysis and link visualization, allowing supply chain managers to have a clearer understanding of the operation of the entire supply chain~\cite{wamba2023both,jayasundara39revolutionizing}.

\subsubsection{Using LSFMs detect malicious perturbations}
Unlike humans, deep learning models might have been more inclined to utilize information such as textures and colors for inference~\cite{geirhos2022imagenettrained}, leading to a heightened sensitivity of deep neural networks towards imperceptible malicious perturbations (referred to as adversarial attacks). The emergence of LSFMs brought promising solutions for this issue\cite{du2023spear}. On one hand, relying on superior generative capabilities, LSFMs could potentially make adversarial attacks more efficient~\cite{barrett2023identifying}, while, on the other hand, LSFMs were more likely to detect hidden malicious perturbations with the superior extraction helps of their capabilities. Meanwhile, inconsistencies between images containing imperceptible perturbations and diffusion-reconstructed images could be identified using diffusion models, thereby recognizing the malicious attacks concealed within input samples~\cite{cao2023impress}. LLMs, a self defense mechanism utilized the LLM output content as re-input, could be used to inspect whether the output content was harmful~\cite{phute2023llm}, randomly dropping a portion of the input and aggregating outputs from various dropping could be used collectively to determine if a request was malicious~\cite{cao2023defending}. In practice, Mireshghallah et al.~\cite{xu2021privacy} combined homomorphic encryption with LSFMs, to allow model inference on encrypted data, thus offering further data protection.

\input{Table/LSFM_roadmaps}

%% file: Table/LSFM_roadmaps.tex
\begin{table*}[]
\resizebox{1.0\linewidth}{!}{
\begin{tabular}{ccccc}
\hline
\textbf{\begin{tabular}[c]{@{}c@{}}Challenges in the Application of \\ Deep Learning in \\ Intelligent Manufacturing\end{tabular}} & \textbf{Manifestation of challenges}                                                                                                                                                                                     & \textbf{\begin{tabular}[c]{@{}c@{}}The technical advantages of \\ solving this problem with a LSFMs\end{tabular}} & \textbf{Roadmaps for solving problems with LSFMs} & \textbf{Related Works} \\ \hline
\multirow{7}{*}{Poor generalization ability}                                                                                       & \multirow{3}{*}{Overfitting}                                                                                                                                                                                             &                                                                                                                   & Pre-training combined with fine-tuning            & {\cite{hu2021lora,song2023fdalign}}                \\
                                                                                                                                   &                                                                                                                                                                                                                          & \multicolumn{1}{l}{}                                                                                              & Regularization                                    & {\cite{taylor2022galactica,xue2023repeat}}                \\
                                                                                                                                   &                                                                                                                                                                                                                          & \multicolumn{1}{l}{}                                                                                              & Continual Learning/Life-Long Learning             & {\cite{wang2022learning,zhao2023expel,yadav2023exploring}}                \\
                                                                                                                                   & \multirow{2}{*}{Difficulty in representing unstructured data}                                                                                                                                                            & Powerful generalization ability                                                                                   & Building structured data through LSFM             & {\cite{sobhkhiz2023integrating,dosovitskiy2020image,oliveira2023new}}                \\
                                                                                                                                   &                                                                                                                                                                                                                          & \multicolumn{1}{l}{}                                                                                              & Using multimodal LSFMs                            & {\cite{chen2022visualgpt,imagebind,li2023multimodal,garza2023timegpt}}                \\
                                                                                                                                   & \multirow{2}{*}{\begin{tabular}[c]{@{}c@{}}Lack of pathways to efficiently \\ integrate domain knowledge\end{tabular}}                                                                                                   & \multicolumn{1}{l}{}                                                                                              & Embedding knowledge through prompts               & {\cite{wang2021knowledge,xia2023unsupervised,zhao2023expel}}                \\
                                                                                                                                   &                                                                                                                                                                                                                          &                                                                                                                   & LSFM assisted construction of knowledge graph     & {\cite{meyer2023llm,chen2020review}}                \\ \hline
\multirow{4}{*}{\begin{tabular}[c]{@{}c@{}}Limited high-quality\\ training dataset\end{tabular}}                                   & \multirow{4}{*}{\begin{tabular}[c]{@{}c@{}}Limited labeling Reources, Variety and complexity,\\ Imbalance and rarity, Specificity and uniqueness,\\ Anonymity and privacy concerns,\\ Dynamic environments\end{tabular}} & \multirow{4}{*}{\begin{tabular}[c]{@{}c@{}}Automatic high-quality\\ training dataset generation\end{tabular}}     & Generation of higher-quality datasets             & {\cite{rombach2022high,trabucco2023effective,wang2023score}}                \\
                                                                                                                                   &                                                                                                                                                                                                                          &                                                                                                                   & Improving data quality                            & {\cite{blip,jain2023llmassisted}}                \\
                                                                                                                                   &                                                                                                                                                                                                                          &                                                                                                                   & Zero-shot and few-shot                            & {\cite{gu2023anomalygpt,leite2023detecting}}                \\
                                                                                                                                   &                                                                                                                                                                                                                          &                                                                                                                   & Pre-training combined with fine-tuning            & {\cite{brown2020language,gao2023desam,radford2019language}}                \\ \hline
\multirow{10}{*}{Unsatisfactory performance}                                                                                       & \multirow{3}{*}{Unsatisfactory accuracy}                                                                                                                                                                                 &                                                                                                                   & Improvement by prompts                            & {\cite{ji2023segment,deng2023samu,dai2023samaug}}                \\
                                                                                                                                   &                                                                                                                                                                                                                          & \multicolumn{1}{l}{}                                                                                              & Enhanced input data                               & {\cite{huang2023robustness,ji2023sam,tang2023can,sun2015learning,aslam2022removal,huang2019image,fan2021concealed,he2023camouflaged}}                \\
                                                                                                                                   &                                                                                                                                                                                                                          &                                                                                                                   & Pre-training combined with fine-tuning            & {\cite{brown2020language,han2022survey,hu2023skinsam,chu2023finetune,DBLP:journals/corr/abs-2110-07602,hu2021lora,dettmers2023qlora}}                \\
                                                                                                                                   & Single task, single modal                                                                                                                                                                                                &                                                                                                                   & Cross modal pre-training                          & {\cite{gu2022wukong,xie2023ccmb,zhou2023rc3,li2023enhanced}}                \\
                                                                                                                                   & \multirow{3}{*}{Lack of interpretability}                                                                                                                                                                                & Superior performance                                                                                              & Using LSFM’s own output for interpretation:       & {\cite{bubeck2023sparks,9658550}}                \\
                                                                                                                                   &                                                                                                                                                                                                                          & \multicolumn{1}{l}{}                                                                                              & Using LLM to explain other models                 & {\cite{singh2023explaining,gat2023faithful,singh2022augmenting}}                \\
                                                                                                                                   &                                                                                                                                                                                                                          & \multicolumn{1}{l}{}                                                                                              & Visualize the running process                     & {\cite{chefer2021transformer,singh2023explainingdata,wamba2023both,jayasundara39revolutionizing}}                \\
                                                                                                                                   & \multirow{3}{*}{Data security}                                                                                                                                                                                           & \multicolumn{1}{l}{}                                                                                              & Pre-training combined with fine-tuning            & {\cite{abadi2016deep}}                \\
                                                                                                                                   &                                                                                                                                                                                                                          & \multicolumn{1}{l}{}                                                                                              & Adopting distributed learning                     & {\cite{fan2023fatellm,kuang2023federatedscope}}                \\
                                                                                                                                   &                                                                                                                                                                                                                          & \multicolumn{1}{l}{}                                                                                              & Using LSFMs for encryption                        & {\cite{du2023spear,cao2023impress,phute2023llm,cao2023defending,xu2021privacy}}                \\ \hline
\end{tabular}
}
\caption{Roadmaps for solving problems with LSFM}
\label{tab:LSFM_roadmaps}
\end{table*}

%% file: Main_text/V_Cases_of_application_of_LSFMs_in_intelligent_manufacturing.tex
\section{Cases of application of LSFMs in intelligent manufacturing}
\label{Cases}

Eventhough LSFMs have intrinsic advantages in terms of powerful generalizations, abilities to establish high-quality training datasets automatically, and superior performance, it is not straight forward to employ them in intelligent manufacturing scenarios. Researchers need to work closely with big manufacturing companies where large-scale industrial data are available to build training datasets and in-situ LSFMs can be tested in real manufacturing lines. Moreover, issues like transferabilities, costs, real-time performance, etc., also require careful investigation before LSFMs can be widely used in large-scale intelligent manufacturing scenes.

In this section, we elucidated a couple of real applications of LSFMs in manufacturing lines of Midea Group to show how LSFMs could help industries improve their efficiency and cut down their cost.

\subsection{Case: PCB defect inspection}
As an implementation form of intelligent manufacturing, industrial intelligent detection technology based on multimodal large models, especially in the field of defect detection in high-density electronic components, is gradually becoming a research hotspot in the industrial production field. This technology can effectively integrate and analyze various types of data, such as visual images, text information, point cloud data, etc., to achieve comprehensive perception and in-depth understanding of electronic devices. On this basis, by applying complex algorithms and computational models, the intelligent detection system can extract and fuse multimodal features, optimize production processes, detect equipment failures, ensure product quality, and ultimately achieve the goal of cost reduction and efficiency improvement. In addition, this technology can also support rapid product switching and personalized customization, improving the responsiveness and flexibility of enterprises to market changes.

Traditional methods only achieved a recognition accuracy rate of around 80\%, unsatisfactory for real manufacturing requirements. Furthermore, direct applications of SAM could introduce issues, such as unlabelled segmented targets,segmentation errors,missed segmentations,under-segmentation and over-segmentation. In reality, we significantly enhance the accuracy of automatic component position detection to 97\% by freezing the image encoder layer and fine-tuning with LoRA. This improvement had the potential to reduce 2000 production line employees, i.e. an annual cost reduction of 150 million yuan for Midea Group.

\begin{figure*}
    \centering
    \includegraphics[width=1\linewidth]{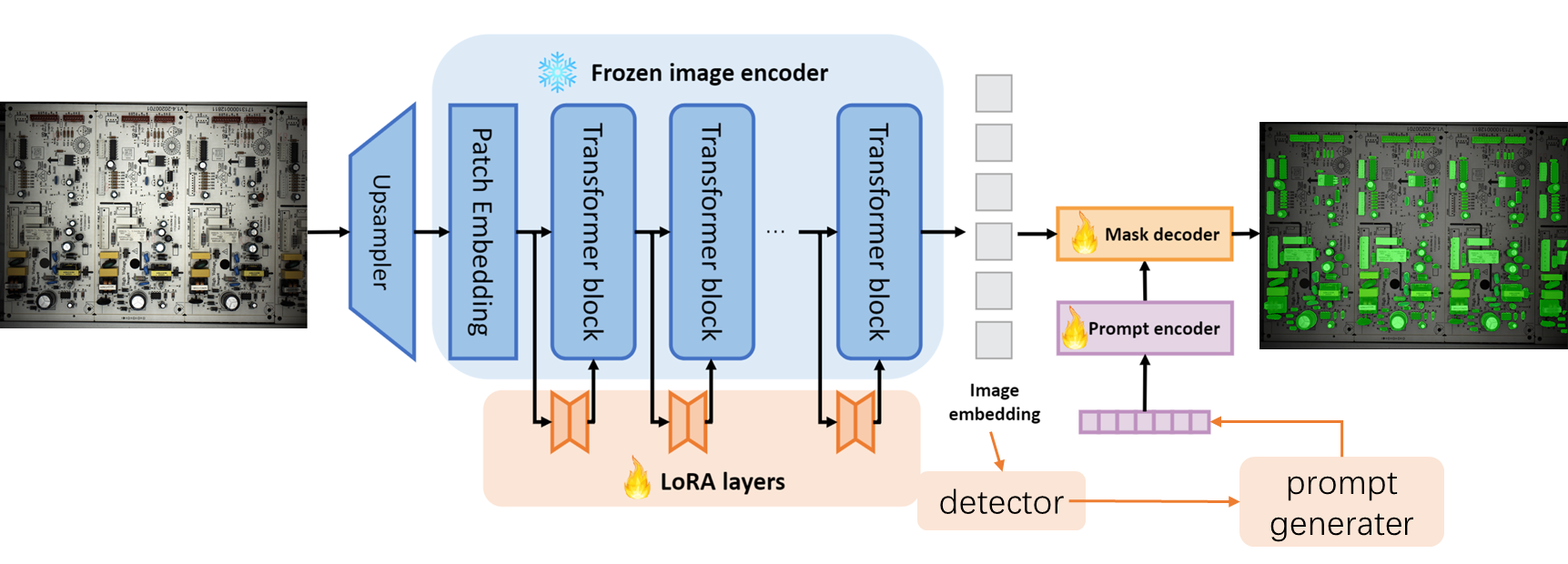}
    \caption{Framework of PCB defect inspection in industrial production}
    \captionsetup{}
    \label{fig:case1}
\end{figure*}

\subsection{Case: Industrial human action recognition}
With the rapid development of intelligent manufacturing, industrial human action recognitions (IHAR) drew great attentions\cite{sun2022human} and started to play important roles in many industrial scenarios, such as quality control\cite{yan2022yolo}, safety monitoring\cite{aiello2022worker}, and process optimization\cite{chen2020discrete}. Nevertheless, as IHAR scale increased in practice, the growing diversity of industrial actions
involved in manufacture lines (MLs) raised complexity of data distribution and transferability across different MLs. 

Constructing a generic industrial dataset for IHAR model training was challenging or even unrealistic\cite{ahmad2022deep}, and up until now, manual annotations were still a mainstream approach for obtaining high quality industrial data, especially when supervised deep learning(DL) models were used. These made it very costly to employ conventional DL-Based IHAR methods, especially when dealing with large-scale industrial data.

In this work, we proposed a LSFM-based low-cost and real-time industrial human action recognition (LRIHAR) model, with the least human intervention. Grounding DINO\cite{liu2023grounding} and BLIP2\cite{5li2023blip} were used for action detection and recognition of large-scale industrial human actions, respectively. After obtaining sufficient boxed action pictures, YOLOv5 was trained as the detector for real deployment. In order to get high accuracy, we trained ViT-L for classification using low-rank adaptation (LoRA) fine-tuning method\cite{hu2021lora}. Finally, knowledge distillation (KD)\cite{duval2023simple} was used to distill low algorithm call time and highly generalized ViT-S models to classify actions selected from YOLO. To summarize, our main contributions were:

(1)Grounding DINO and BLIP2 was jointly used to facilitate automatic annotations and industrial dataset establishment, wherein annotation costs was saved by more than 80\% with superior generalization over traditional meth-
ods, speeding up training process and IHAR deployment.

(2)LoRA and KD was used to reduce time for training and response, respectively, wherein 96.84\% classification accuracy was achieved, outperforming retrained ResNet-18 in all scenarios with call time less than 10ms, realizing
real-time IHAR.

(3)Comprehensive experiments on large scale industrial data from three industrial MLs of Midea Group showed that the LRIHAR achieved 98.19\% detection accuracy on average, and surpassed the conventional ResNet-18 by 4.4\% on classification accuracy. Moreover, the LRIHAR showed superior AC cost saving, real-time performance, and generalization capabilities, simultaneously, facilitating its applications in large scale industrial scenarios.

\begin{figure*}
    \centering
    \includegraphics[width=1\linewidth]{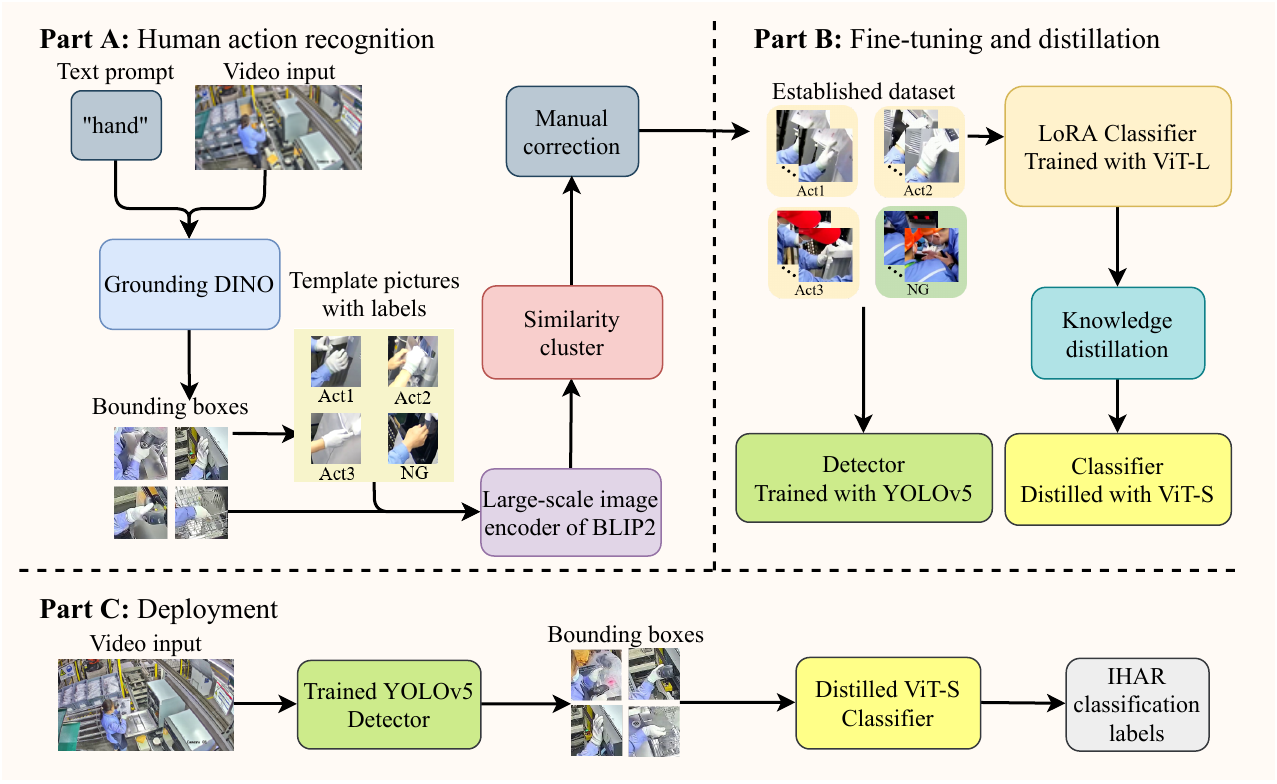}
    \caption{Framework of real-time industrial action recognition on industrial production lines}
    \captionsetup{}
    \label{fig:case2}
\end{figure*}

%% file: Main_text/VI_Conclusion.tex
\section{Conclusion}
LSFMs demonstrated powerful generalization capabilities, the ability to automatically generate high-quality training datasets, and superior performance, and was able to transform AI from a paradigm of single modal, single task, and training on limited data to a pattern of multimodal, multitask, and pre-training on massive data followed by fine-tuning, and were bound to bring about a new wave of transformation in intelligent manufacturing.

Since research on applying LSFMs on intelligent manufacturing was still in its early stages and it was lack of systematic directional guidance in this areas, this paper summarized the progresses and challenges of deep learning in intelligent manufacturing, as well as the progresses of LSFMs and their potential advantages in intelligent manufacturing applications. Accordingly, this paper comprehensively discussed how to construct an LSFM system suitable for the intelligent manufacturing domain from the perspectives of generalization, data, and performance, and illustrated how the application of LSFMs could help enterprises enhance efficiency and reduce costs by presenting the real-world applications of LSFMs in the manufacturing lines of Midea Group.